\documentclass{article}
\usepackage{arxiv}
\usepackage{cite}
\usepackage{amsmath,amssymb,amsfonts}
\usepackage{algorithm}
\usepackage{algorithmicx}
\usepackage{algpseudocode}
\usepackage{graphicx}
\usepackage{textcomp}

\usepackage{soul}
\usepackage{setspace}
\usepackage{hyperref}
\usepackage{subcaption}

\usepackage{epsfig}
\usepackage{multirow}
\usepackage{lipsum}
\def\BibTeX{{\rm B\kern-.05em{\sc i\kern-.025em b}\kern-.08em
    T\kern-.1667em\lower.7ex\hbox{E}\kern-.125emX}}

\begin{document}

\title{Human Action Performance using Deep Neuro-Fuzzy Recurrent Attention Model}

\author{  Nihar Bendre , Nima Ebadi \thanks{The first two authors have equal contribution} \\
  Secure AI \& Autonomy Laboratory\\
  Department of Electrical and Computer Engineering\\
  University of Texas at San Antonio \\ San Antonio, TX 78249, USA
  \And
  John J Prevost  \\
  Department of Electrical and Computer Engineering\\
  University of Texas at San Antonio\\ San Antonio, TX 78249, USA
  \And
  Paul Rad  \thanks{Corresponding author: paul.rad@utsa.edu}\\
   Secure AI and Autonomy Laboratory\\
  Department of Electrical and Computer Engineering\\
  Department of Information Systems and Cyber Security\\
  University of Texas at San Antonio\\ San Antonio, TX 78249, USA
  }




\maketitle

\begin{abstract}
\label{abstract}
A great number of computer vision publications have focused on distinguishing between human action recognition and classification rather than the intensity of actions performed. Indexing the intensity which determines the performance of human actions is a challenging task due to the uncertainty and information deficiency that exists in the video inputs. To remedy this uncertainty, in this paper we coupled fuzzy logic rules with the neural-based action recognition model to rate the intensity of a human action as \textit{intense} or \textit{mild}. 
In our approach, we used a Spatio-Temporal LSTM to generate the weights of the fuzzy-logic model, and then demonstrate through experiments that indexing of the action intensity is possible.
We analyzed the integrated model by applying it to videos of human actions with different action intensities and were able to achieve an accuracy of 89.16\% on our intensity indexing generated dataset. The integrated model demonstrates the ability of a neuro-fuzzy inference module to effectively estimate the intensity index of human actions.

\end{abstract}


\section{Introduction}
\label{intro}

Recently, action recognition based on supervised deep learning has attracted a lot of interest in the computer vision research community due to its numerous applications in video analytics, surveillance, security, sports analysis, and human-computer-interaction based applications \cite{kong2018human}. Researchers all over the world are doing extensive studies on various techniques to propose models with better performance \cite{singh2017online,luvizon20182d,rahmani2018learning}. Despite these efforts, this field still poses many challenges which include intra-class variation, viewpoint orientation, occlusion, various motion speed and different styles of background clutter. A drawback to the supervised deep learning approach of action recognition is that less focus is given to predict the intensity of the action \cite{mabrouk2018abnormal, rapantzikos2009dense, kay2017kinetics, mavadati2013disfa}. Determining the intensity of an action is crucial in environments like bullying and violence detection in school, at work, at home, in public areas, and in prison \cite{patwardhan2016aggressive,dankert2009automated, silva2019cooperative, silva2017multi}. Intensity indexing can also be used for detecting aggressive behavior in applied behavior analysis (ABA) \cite{bailey2017research}, a proven assessment and treatment model for Autism Spectrum Disorder (ASD) \cite{dillenburger2009none} and other severe mental disorders \cite{harvey2009application}. In the context of ASD, intensity index can aid caretakers in assessing danger in patients' behavior and prevent serious health consequences such as concussion from head banging \cite{soke2016brief}.

Action intensity index is defined as a measure of kinetic intensity used to determine whether a specific action is performed with high or low intensity. Kinetic intensity is the amount of kinetic power it takes to perform a certain action, and can be applied to the concept of indexing intensity of human actions \cite{van2002rainfall, chun2016human}. The kinetic power of a certain action is directly proportional to the velocity and the mass of the moving object \cite{van2002rainfall}. However, in the context of human activities, which involve the movement of human joints, the kinetic power depends on the velocity of the joints engaged in the main activity \cite{shan20143d}, as well as the number and extent in which they are engaged \cite{zhang2015adaptive,chun2016human}: more moving joints utilizing more joint power results in greater kinetic power and intensity.\footnote{Note: Action intensity indexing is different than velocity and includes extreme activities with high kinetic power.}

\begin{figure*}[t!]
\small
\begin{subfigure}{.25\linewidth}
  \centering
  \includegraphics[width=0.9\linewidth]{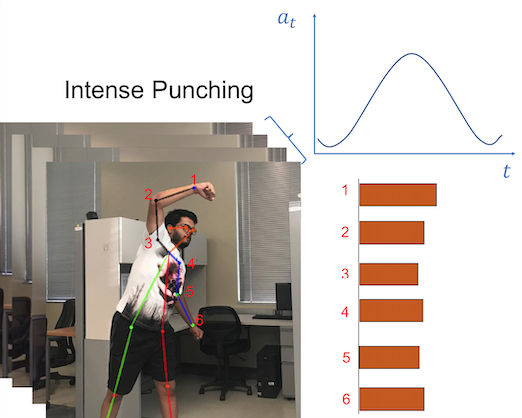}
  \caption{\small{Subject A Punching Hard}}
  \label{fig:nihar_hitting_hard}
\end{subfigure}%
\begin{subfigure}{.25\linewidth}
  \centering
  \includegraphics[width=\linewidth]{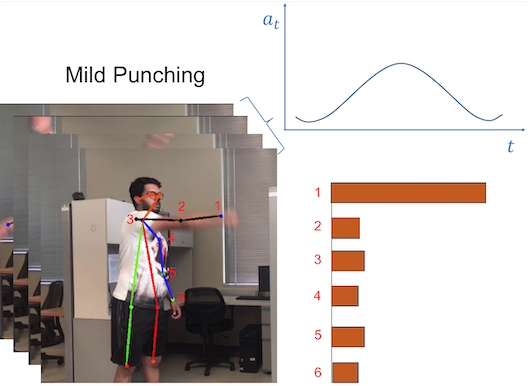}
  \caption{\small{Subject A Punching Soft}}
  \label{fig:nihar_hitting_soft}
\end{subfigure}
\begin{subfigure}{.25\linewidth}
  \centering
  \includegraphics[width=0.9\linewidth]{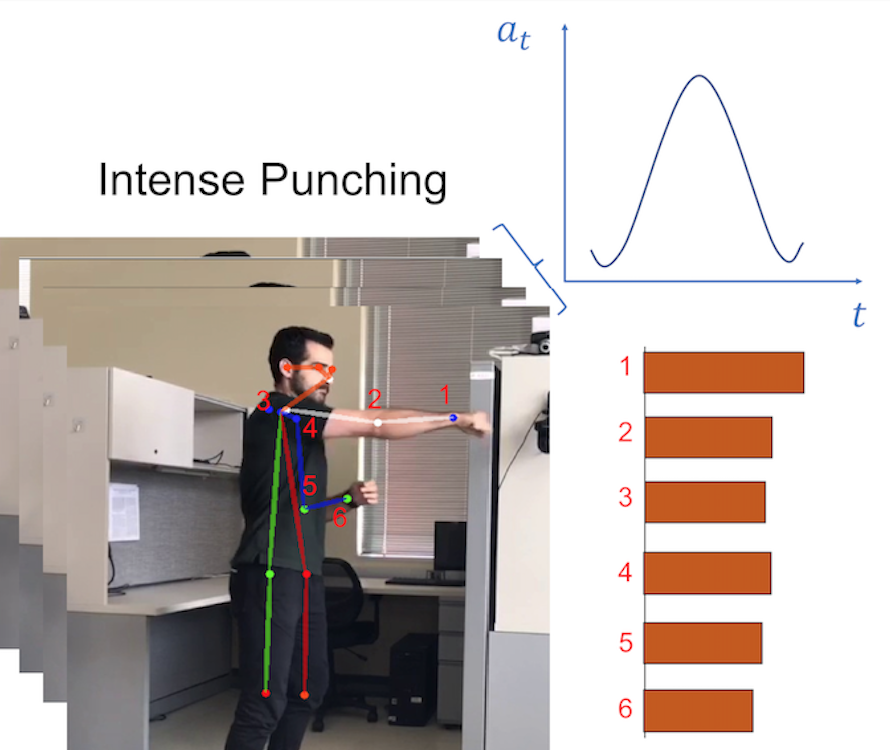}
  \caption{\small{Subject B Punching Hard}}
  \label{fig:mehrad_hitting_hard}
\end{subfigure}%
\begin{subfigure}{.25\linewidth}
  \centering
  \includegraphics[width=\linewidth]{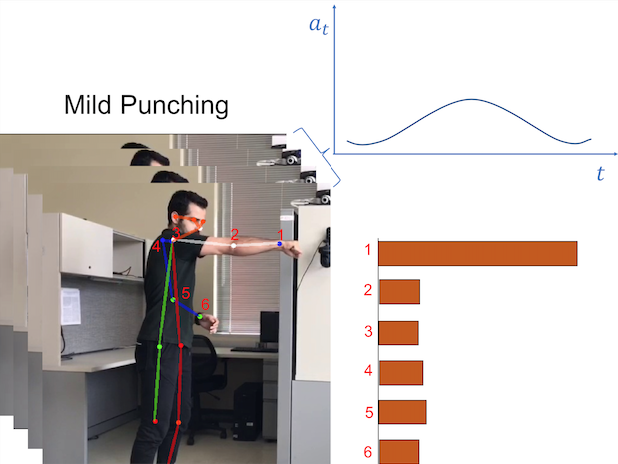}
  \caption{\small{Subject B Punching Soft}}
  \label{fig:mehrad_hitting_soft}
\end{subfigure}%
\caption{\small{The distribution of attention weights (over joints and time) changes with the intensity index of performed action. The line plot indicates attention over time frames and the bar plot indicates attention over the joint movements. Even though both subjects are performing the same action, the distributions of attention weights are different}}
\label{fig:first_Page}
\end{figure*}
The intensity of human actions cannot be generalized into a single, crisp formula as it varies from person to person. Intensity is rather a subjective term in which some level of uncertainty is always present, often expressed using imprecise language. Furthermore, measurement inaccuracies are inevitable from a 2D video. Therefore, to measure the intensity of an action from an input video, a mathematical model is required which accounts for such uncertainties and inaccuracies by modelling and minimizing their effects. \cite{mendel2017uncertain}.

While deep learning based models can help with learning adaptation and scaling up to more general applications \cite{lecun2015deep}, they cannot capture data or model uncertainty \cite{gal2016dropout}. In addition, deep learning based models lack the human-like ability to interpret imprecise information. Fuzzy inference systems, on the other hand, provide an inference mechanism for uncertainty and enable the qualitative interpretation of the actions. In the context of intensity indexing, deep learning based models also encounter serious problem when the dataset is biased towards a specific way of performing an action. These models are not able to learn dissimilarities in human motions when actions are performed with various intensities. However, adaptive fuzzy systems can generate membership functions for different types of target action intensities \cite{yager2012introduction}. To enable our system to deal with the uncertainty and varied nature inherent in this application, we propose a hybrid system combining the concept of fuzzy logic and deep recurrent neural networks. Such integration has proven effective in a wide variety of real-world problems \cite{li2019unified,tran2018closer,bonanno2017approach}.

Our proposed methodology is an attentive neuro-fuzzy system designed to recognize qualitative differences in human actions and to self-adapt to different intensities. Inspired by the model proposed by \cite{rutkowski2003flexible}, our model utilizes recurrent neural networks to detect actions from spatio-temporal patterns of human poses, in tandem with an adaptive fuzzy inference system to learn the various human motions used to perform actions with different intensities and then estimate the action's intensity.

The integrated model can successfully learn the unique way a specific action with a certain intensity is performed, as well as estimate the intensity of the respective action. Experimental results prove the effectiveness of the integrated model in recognizing the action movements of different intensities. To the best of our knowledge, our framework is the first to index the intensity of action from an input video. Our contributions in this paper are:
\begin{itemize}
\item We propose a novel hybrid model based on a fuzzy inference system coupled with a spatio-temporal LSTM action recognition module to jointly determine the intensity index of the recognized action.
\item Our work provides a case study on a generated dataset of human actions with two intensity indexes: \textit{intense} and \textit{mild}, to evaluate the performance of our model in more fine-grained recognition of actions and intensities.
\end{itemize}

\begin{figure*}[ht!]
\small
\includegraphics[width=0.9\linewidth]{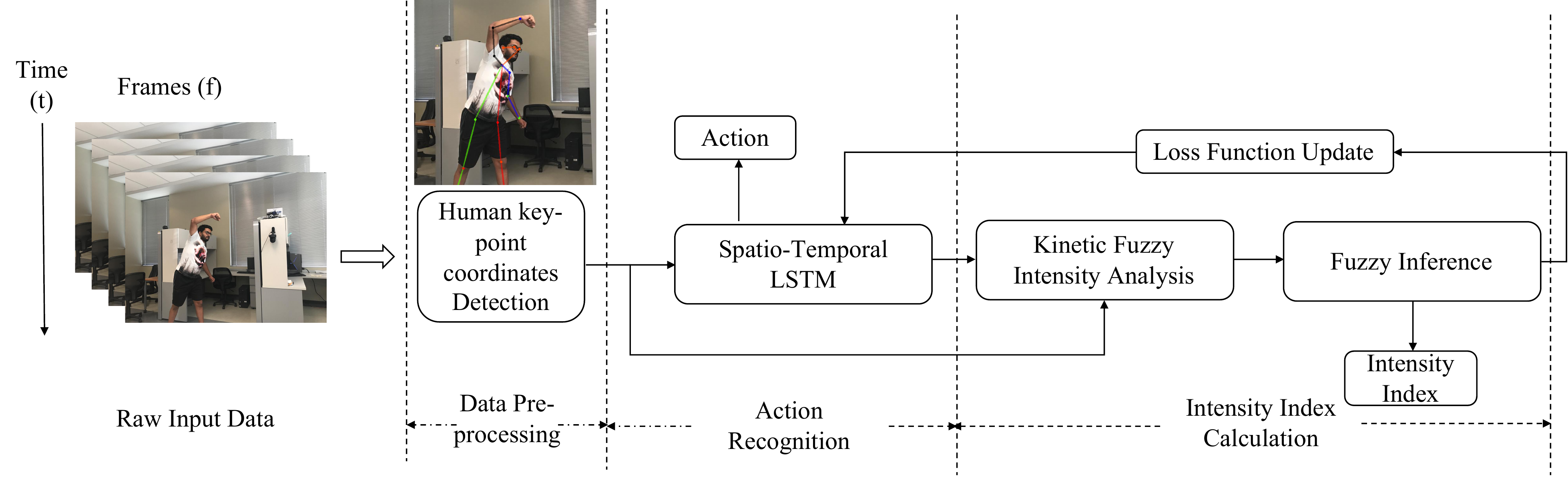}
\caption{\small{The camera frames act as an input to the data pre-processing stage where the human key-point coordinates are generated. These coordinates act as an input to the Spatio-Temporal LSTM, which detects the actual action and generates the attention weights. These weights are the input to the Kinetic Fuzzy Intensity Analysis, which generates the intensity score. This intensity score dynamically updates the fuzzy logic rules and is also used to determines the Intensity Index of the performed action.}}
\label{fig:overview}
\end{figure*}

\section{Related Work}
\label{Related_Work}
The related work for the model is based on the deep learning components which are used for data pre-processing (as discussed in methodology section of this paper) and to train the model on action recognition based on supervised learning. Our model leverages deep learning components as well as neuro-fuzzy systems to dynamically generate fuzzy logic rules to detect the intensity of various human actions. To detect actions using human key-point coordinates, the model requires spatio-temporal information of the scene so it can also be seen as a time series problem. To overcome this, much research has been done in the recent past to come up with models that can effectively predict actions from key-point coordinates \cite{presti20163d}. Traditional methods involved features which were hand crafted to represent the inter frame relationship of the key-point coordinates sequence \cite{simonyan2014two, de2018driverless, de2019implementation,das2019distributed, wang2013action, qiu2017learning}. Recent studies have utilized deep learning techniques to detect and predict relationships by using spatio-temporal information in a collection of frames. \cite{minh2018fine} designed a Fine-to-Coarse Deep Convolutional Neural Network (CNN) along with fully connected layers which extract the spatio-temporal and spatial features of a key-point coordinates sequence. Furthermore, the use of 3D-CNN with a 3D filter kernel has also been proved to be able to learn the spatio-temporal information \cite{qiu2017learning}. To capture temporal information, research has been done to predict action using Recurrent Neural Networks (RNN) which are based on Long Short-Term Memory (LSTM) or Attention Models. There have been few research examples where a RNN based model has been used to predict the actions from a human key-point coordinates \cite{liu2016spatio,liu2017skeleton,song2017end,zhu2016co}. Recent research also points to the use of a Convolutional-Recurrent Neural Network (CRNN) where the CNN is used to extract the features from the input frames and the output of the CNN is fed to LSTM to extract the temporal dynamics. State-of-the-art results were achieved with the use of a graphical neural network \cite{li2018action,yan2018spatial}. Compared to the graphical neural network and LSTM, CNN displays better results for learning to represent images in terms of key-point coordinates representation \cite{ke2017new,kim2017interpretable}, but their performance drops when dealing with long spatio-temporal sequences.

While deep learning models can achieve better scalability and can generalize better, they lack in capturing data uncertainty, subjectivity and human-like reasoning \cite{lecun2015deep}. Fuzzy logic can capture the uncertainty, subjectivity and have human-like reasoning \cite{zadeh1965fuzzy}. The proposed idea is to use fuzzy inference on top of a deep learning action recognition module to index intensity of the action as either mild or intense \cite{mabrouk2018abnormal, rapantzikos2009dense, kay2017kinetics, mavadati2013disfa}. Indexing the intensity of a subjective task and involves a certain amount of uncertainty from individual to individual. It requires adaptive learning which cannot be derived by just stacking various modules sequentially.

\section{Methodology}
\label{Methodology}
This paper proposes a novel neuro-fuzzy system using recurrent neural networks and fuzzy inference systems which adaptively perform fine-grained recognition of human action intensity indexes. As shown in \autoref{fig:overview}, our methodology consists of three processing sections: data preprocessing, action recognition, and intensity indexing. First, the preprocessing section transforms the input video of an action to a tensor of the human key-point coordinates over time using a pose detection algorithm. This tensor is next passed to an LSTM network to recognize the human action based on the spatio-temporal patterns existing in the tensor. The LSTM model is equipped with two self-attention mechanisms \cite{vaswani2017attention}, one over the time frames and another over the coordinates. The attention weights, along with the coordinate's tensor, are then fed to the kinetic fuzzy intensity analysis module. The kinetic fuzzy intensity analysis module computes an initial intensity score based on fuzzy entropy measures. The fuzzy inference module converts the intensity score and the attention weights into fuzzy sets using an adaptive membership function. Using the truth values of these fuzzy sets, our methodology defines the fuzzy rules through which the final intensity index is determined. Finally, the spatio-temporal LSTM's loss function gets updated with a customized penalty term to further adapt to distinct movements of intense-mild actions.

\subsection{Data Pre-processing}
\label{Data_Pre-processing}
Before the raw data can be input int the action recognition module, human key-point coordinates must be generated using the pose estimation technique. Using of human key-point coordinates to train the action recognition module will help to reduce the background clutter \cite{liu2016spatio,liu2017skeleton}. Also, it will reduce the computational complexity as compared to using the entire image/video to train the module \cite{song2017end,zhu2016co}. We also need to feed the human key-point coordinates to the neuro-fuzzy section for qualitative action recognition. To extract the human key-point coordinates we use the model described by \cite{wei2016cpm,cao2017realtime,simon2017hand} which achieves, state-of-the-art results on multiple public benchmarks for pose estimation and human key-point detection.

\begin{figure*}
\small
\includegraphics[width=\linewidth]{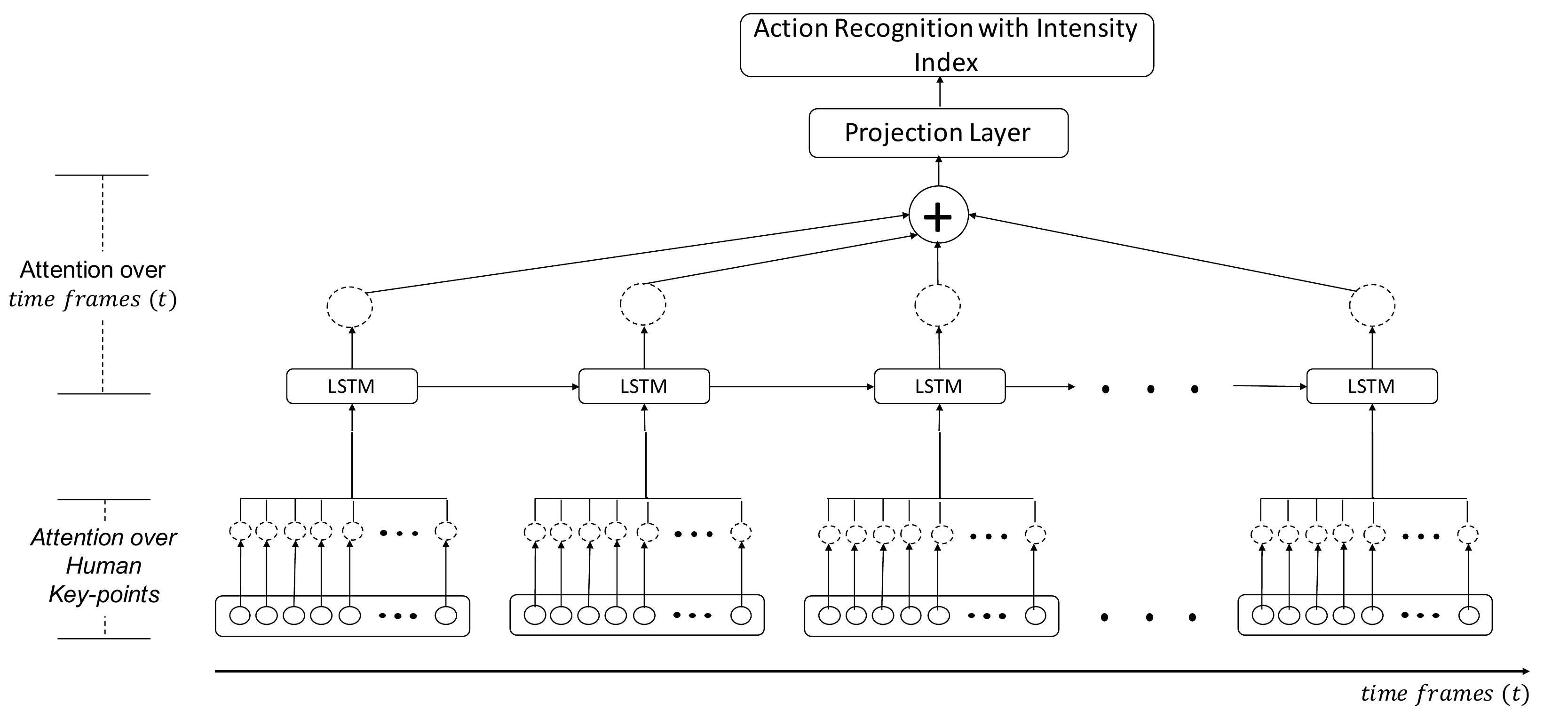}
\caption{The attention over human key-points defines the parts of the human bodies involved in the action can be observed from and the attention over time frames defines the frames in which the action is being performed. The attention weights also show the significance of the respective frames in recognition of the action.}
\label{fig:attention}
\end{figure*}

\subsection{LSTM with Spatio-Temporal Attention}
\label{section:LSTM}
To capture the sequential patterns of key point coordinates, we utilize LSTM, an RNN \cite{hochreiter1997long,das2018distributed,hochreiter2001gradient}. For supervised deep learning, there are various LSTM models developed for action detection \cite{li2015category,chacon_poison,sahba2018image,zhu2016co,ke2017new}. Inspired by Liu's (2016) Spatio-Temporal LSTM model, we apply a similar model that is equipped with a spatio-temporal mechanism to recognize the performed action and learn the exclusive motion patterns of the action. In other words, our inspired LSTM model utilizes two attention mechanisms \cite{sahba2018automatic}: attention over the time frames, and attention over various key-point coordinates. Such spatio-temporal attention helps the model to understand an action despite variation among individuals preforming the same action with a certain intensity index, such as walking fast or punching hard. One attention mechanism is implemented on top of the recurrent architecture of the LSTM cells, and the other one is implemented across the units of input and hidden states, so that the model can selectively focus on the time frames as well as human key-point coordinates (see \autoref{fig:attention}). These two attention mechanisms demonstrate the engagement of the human key-point coordinates in each time frame in the detected action. In addition to learning the possible behavioral variation of performing an action, the weights of these two attention mechanisms are used to measure the kinetic intensity score and determine the fuzzy inference of the intensity index.

\begin{figure*}
    \centering
    \small
    \includegraphics[width=\linewidth]{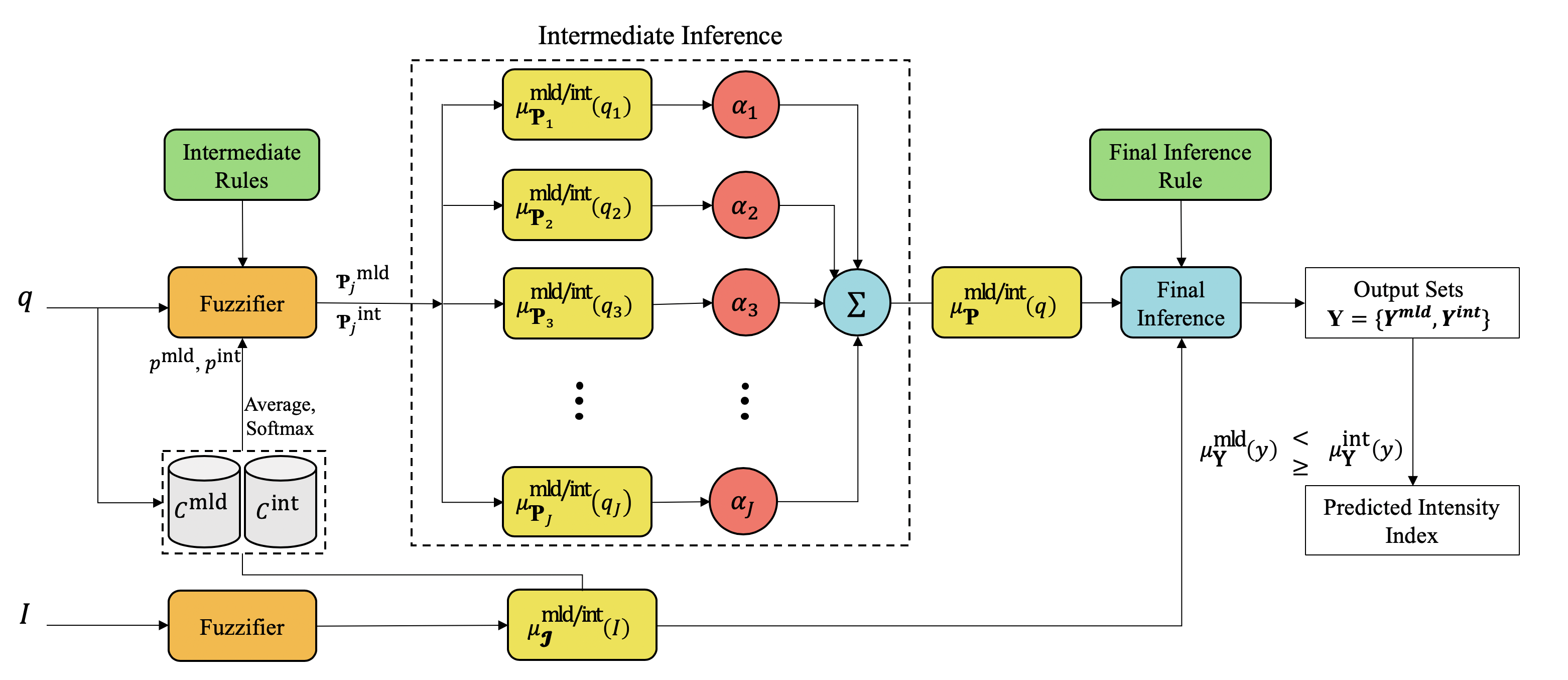}
    \caption{The procedure for our fuzzy inference system. The crisp input values are mapped into fuzzy sets through fuzzification processes, i.e. fuzzifiers. Next, through intermediate rules and a linear combination on fuzzy inputs, an intermediate inference is computed. The linear combination uses adaptively trained weights, i.e. adaptive filter. The final inference is performed on the intermediate output set and the fuzzified intensity score flowing through the final inference rules.}
    \label{fig:fis}
\end{figure*}

\subsection{Kinetic Intensity Score using Fuzzy Entropy}
\label{kinetic}
Once the attention LSTM model is trained to recognize the performed action, we utilize the parameters of the attention vectors along with fuzzy entropy measures to compute an initial intensity score for an instance of the action. This initial kinetic intensity score is utilized to generates dynamically fuzzy rules to specify the index of the intensity as \textit{intense} or \textit{mild}. As shown in \autoref{fig:attention}, the Spatio-Temporal LSTM model is equipped with a self-attention mechanism \cite{vaswani2017attention,ebadi2019implicit} that detects the time frames in which the detected action is happening, extracting a linear combination of the hidden states to the output and generate the temporal attention weights. These weights denote the amount of influence of each time frame in the final inference. In other words, they determine whether, and to what confidence level, an action is observed in each time frame. We can utilize the distribution of these weights to measure the intensity of the action. For instance, the faster an action happens, the observation of the performed action in time, is less and resultant distribution of temporal attention weights is proportionally denser. Therefore, the entropy of these attention weights has an inverse relationship with the intensity and speed of an action. 

The intensity of action depends on the kinetic energy of the limbs that are engaged in performing the action which are translated into key-point coordinates. Shan et al., \cite{shan20143d} formulate this kinetic energy by the movement of the key-point coordinates over the video frames. Thus, we consider this kinetic energy by by adding it to the attention distribution as fuzzy membership weights and computing their fuzzy entropy. The weights are the change of the coordinates' locations from the last frame multiplied by their corresponding attention weights, following \cite{shan20143d}. Using fuzzy entropy methods from \cite{shannon1948mathematical,zadeh1965fuzzy} , we calculate the fuzzy entropy of the attention vector which is \textit{indirectly} related to intensity, as follows:

\begin{equation}
H_{fuzzy}(a_t, \mu_t) = -\sum_{t=1}^{T} a_t \cdot \mu_t \cdot \log({a_t}) \end{equation}

\begin{equation}
\mu_t = \frac{1}{(a_t \cdot |\Delta x_t|)} = \frac{1}{(a_t \cdot |x_t - x_{t-1}|)}
\end{equation}

\begin{equation}
\label{equ:fuzzy_1}
\implies H_{fuzzy}(a_t, \Delta x_t) = -\sum_{t=1}^{T}\frac{\log ({a_t})}{|x_t - x_{t-1}|}
\end{equation}

where, $x_t$ is the input at the time frame $t$, $\mu_t$ is the membership weight for $H_{fuzzy}(a_t, \mu_t)$ at time $t$, and $a_t$ is the attention weight over time frame $t$.

Furthermore, the intensity of an action also depends on the number of the engaged joints \cite{chun2016human}. As a concrete example, an intense punch, in comparison to a mild one, includes the movement of a greater number of joints across more dimensions such as hip rotation and non-dominant hand movement. Zhang et al. \cite{zhang2015adaptive} use the same concept to quantify the intensity of human facial actions by the number of engaged coordinates and how much they are engaged. As such, we also extract multi-dimensional attention over the human key-points coordinates. Just as we calculated the fuzzy entropy of the directional attention, we also calculate the fuzzy entropy of the dimensional attention which is \textit{directly} related to intensity. The fuzzy weights are the product of temporal attention and dimensional attention over every time frame.

\begin{equation}
\label{equ:fuzzy_entropy}
H^{\prime}_{fuzzy}(a^{\prime}_{(j,t)},a_t) = -\sum_{t=1}^{T} a_t \cdot \sum_{j=1}^{J} a^{\prime}_{(j,t)} \cdot \log( {a^{\prime}_{(j,t)}})
\end{equation}

where $a_t$, attention weight over time frame $t$, is the fuzzy weight. $a^{\prime}_{(j,t)}$  is the attention weights over the key-point coordinate (i.e. human joints) at time frame $t$.

Finally, considering both kinetic energies which has been similarly used in the literature for action intensity \cite{shan20143d,zhang2015adaptive,chun2016human}, we formulate intensity as the proportion of the fuzzy entropy in \autoref{equ:fuzzy_entropy} over the the fuzzy entropy in \autoref{equ:fuzzy_1}. In other words, we measure the kinetic intensity through the fuzzy entropy of the attention weights over the coordinate's locations divided by the fuzzy entropy of the attention weights over the time frames. As follows:

\begin{equation}
I = \frac{H^{\prime}_{fuzzy}(a^{\prime}_{(j,t)},a_t)}{H_{fuzzy}(a_t, |\Delta x_t|)}
\label{equ:fuzzy_entropy_attn}
\end{equation}

where, $I$ is the intensity score, $a^{\prime}_{(j,t)}$ is the attention weight over the key-point joint $j$ at time frame $t$.

\subsection{Fuzzy Inference for Intensity Indexing}
\label{Intensity_Indexing}
As mentioned in \autoref{intro}, intensity is not a very precise term and there is no general formula to measure a crisp value of it. Therefore, after computing the kinetic intensity score, our methodology uses an adaptive fuzzy inference system to detect the intensity index based on both the kinetic intensity score $I$ computed from the previous section, and distribution of the joints' attention weights $q=\left\langle q_j \right\rangle$ as motion patterns. These two values are fed to our fuzzy inference system as crisp input values. This procedure is illustrated in \autoref{fig:fis}.

\subsubsection{Fuzzification of Intensity Score and Joints' Distribution}
\label{sec:::fuzzifier}
In this regard, using dynamically learned membership functions, these crisp input values are mapped to fuzzy sets: $\mathcal{I} = \{\mathcal{I}^{mld},\mathcal{I}^{\text{int}}\}$ and ${\mathcal{P}_j} = \{{\mathcal{P}_j}^{\text{int}},{\mathcal{P}_j}^{mld}\}$\footnote{Generally, the input values can partitioned into $K$ categories/regions, i.e. $\mathcal{I} = \{\mathcal{I}^{0},\mathcal{I}^{\text{1}}, ... \mathcal{I}^{\text{K-1}}\}$ and ${\mathcal{P}_j} = \{{\mathcal{P}_j}^{\text{0}},{\mathcal{P}_j}^{1}, ... , {\mathcal{P}_j}^{K-1}\}$. But in this project, for the sake of simplicity we use $K=2$.}, which denote the partitioning of the intensity score and attention weight corresponding to joint $j$, respectively, into mild and intense regions. Our fuzzy inference system looks at these fuzzy sets as rough estimations of the intensity index. However, the final intensity index output is computed based on these rough estimations and fuzzy logic.

Our model dynamically learns fuzzy membership functions for these fuzzy sets, i.e. $\mu_{\mathcal{I}}$ and $\mu_{\mathcal{P}_j}$, based on the previously computed kinetic intensity score and the distribution of the corresponding attention weights. Using the average intensity index and the common triangular shape, the fuzzy membership $\mu_{\mathcal{I}}$ is formulated as below:

\begin{equation}
\mu_{\mathcal{I}}^{\text{mld/int}}(I) = max(0,0.5+\frac{\mp(I -\Bar{I})}{\sigma})
\label{equ:MFintensityindex}
\end{equation}

where, $\mu_{\mathcal{I}}^{\text{mld/int}}(I)$ refers to the truth values of $\mathcal{I}^{mld}$ and/or $\mathcal{I}^{\text{int}}$ respectively. $\Bar{I}$ is the averaged intensity score, which dynamically gets updated. $\sigma$ defines the spread of the fuzzy set, larger values denote more uncertainty is assumed to exist in the data \cite{mendel2017uncertain}.

The membership function $\mu_{\mathcal{P}_j}$ is adaptively computed using the membership function of \autoref{equ:MFintensityindex} and categorized  distribution of joints' attention weights along with normalized cross-entropy distance. This process is elaborated upon in the followings: Firstly, the model stores the intensity scores of every action as well as the relative attention weights of the spatio-temporal LSTM network. Next, by comparing the truth values of $\mathcal{I}^{mld}$ and $\mathcal{I}^{\text{int}}$ we categorize the stored attention weights into the following categories:

\begin{equation}
\small
C^{mld} = \bigg\{a_{ij}= \frac{ \mu_{\mathcal{I}}^{mld}(I_i)}{T_i} \sum_{t=1}^{T_i} a^{\prime}_{i} (j,t) \bigg| \mu_{\mathcal{I}}^{\text{int}}(I_i) \leq \mu_{\mathcal{I}}^{mld}(I_i)\bigg\}
\label{equ:categorymld}
\end{equation}
\begin{equation}
\small
C^{int} \hspace{0.9mm} = \bigg\{a_{ij}= \frac{\mu_{\mathcal{I}}^{int }(I_i)}{T_i} \sum_{t=1}^{T_i} a^{\prime}_{i} (j,t)  \hspace{0.9mm} \bigg| \mu_{\mathcal{I}}^{int }(I_i) > \mu_{\mathcal{I}}^{mld}(I_i)\bigg\}
\label{equ:categoryint}
\end{equation}

where, $i \in \{1,2,3, ... , N\}$ is the sample index, and $j \in \{1,2,3, ..., J\}$ is the index of human key-point/joint, $N$ is the number of samples and $J$ is the number of joint coordinates, $T_i$ is the number of time frames in the $i^{th}$ sample, $a_{ij}$ is the corresponding joint attention weight averaged over the time frames. 

\begin{algorithm}[t!]
\footnotesize
{
    \caption{The summary of the first stage of our fuzzy inference system which is fuzzification of intensity score $I$ and joints' distribution $q=\langle q_j \rangle$. This process is performed for every input video and updates its values to dynamically adapt to different action intensity indexes.}
    \label{alg:membership_function}
    \begin{algorithmic}
            \State $\Bar{I}$ : average intensity score
            \State ${\Delta}\Bar{H}$ : average difference between the cross-entropy of mild and intense distributions
            \State $C^{int}$ : collection of joint attention weights for intense
            \State $C^{mld}$ : collection of joint attention weights for mild
            \State $p^{int}$ : probabilistic joint distribution for intense
            \State $p^{mld}$ : probabilistic joint distribution for mild
        \For{every input $i$: (${a^i}_t$, ${{a^{\prime}}^i}_{(j,t)}$, $x^i$)}
            
             \Procedure{FUZZIFIER}{${a^i}_t$, ${{a^{\prime}}^i}_{(j,t)}$, $x^i$}
                \State $I^i$ is calculated (Eq. \ref{equ:fuzzy_entropy_attn}), and $\Bar{I}$ is updated (Eq. \ref{equ:MFintensityindex})
                \State truth value $\mu_{\mathcal{I}}^{\text{mld/int}}(I^i)$ is calculated (Eq. \ref{equ:MFintensityindex})
                \State \textproc{*UPDATE-JOINT-DIST} (${{a^{\prime}}^i}_{(j,t)}$, $\mu_{\mathcal{I}}^{\text{mld/int}}(I^i)$) \Comment{Explained at the bottom}
                \State $q^i=\langle {q^i}_j \rangle$ is calculated (Eq. \ref{equ:inputjointsdistibution}) 
                
                \For{every joint $j$}
                    \State $\Delta H$ is calculated between $q_j$ and ${{P}_j}^{\text{mld/int}}$ 
                    \State $\Delta\Bar{H}$ is updated
                    \State $\mu_{\mathcal{P}_j}^{\text{mld/int}}(q_j)$ is calculated (Eq. \ref{equ:MFjointj})
                \EndFor

        \Return $\mu_{\mathcal{P}_j}^{\text{mld/int}}(q_j^i)$\hspace{1mm} $\forall j$ , \hspace{1mm}  $\mu_{\mathcal{I}}^{\text{mld/int}}(I^i)$
        \vspace{0.5mm}
        \EndProcedure
        \EndFor
        
    \end{algorithmic}
    \hrulefill 
    \vspace{0.5mm}
    \begin{algorithmic} 
        \Procedure {*UPDATE-JOINT-DIST}{$a^{\prime}_{(j,t)}$, $\mu_{\mathcal{I}}^{\text{mld/int}}(I)$}
            \If{$\mu_{\mathcal{I}}^{\text{int}}(I) \leq \mu_{\mathcal{I}}^{\text{int}}(I)$}
                \State append $C^{mld}$ (Eq. \ref{equ:categorymld})
                \State update $p^{mld}$ (Eq. \ref{equ:mld/intjointsdistibution})
            \Else
                \State append $C^{int}$ (Eq. \ref{equ:categoryint})
                \State update $p^{int}$ (Eq. \ref{equ:mld/intjointsdistibution}) 
            \EndIf
        \EndProcedure
    \end{algorithmic}
}    
\end{algorithm}

Every weight is multiplied by the corresponding $\mu_{\mathcal{I}}$ to highlight those with higher certainty. Then, by taking the average over various samples of every action, we derive customized distribution of joints' weights for each category of intense-mild. We calculate the softmax of these weights to convert them into probabilistic distributions:

\begin{equation}
p^{\text{mld/int}} = \bigg\{ p^{\text{mld/int}}_j = \footnotesize{\text{softmax}} \bigg(\frac{1}{N} \sum_{i=1}^N a_{ij}\bigg) \bigg| \forall a_{ij} \in C^{\text{mld/int}} \bigg\}
\label{equ:mld/intjointsdistibution}
\end{equation}

Similarly, we derive a probabilistic distribution of joints' weights for every new input sample, as follows:

\begin{equation}
q = \bigg\{ q_j = \footnotesize{\text{softmax}} \bigg( \frac{1}{T_{N+1}} \sum_{t=1}^{T_{N+1}} a^{\prime}_{(j,t)} \bigg) \bigg|  \forall j \in \{1,2,3, ..., J\} \bigg\}
\label{equ:inputjointsdistibution}
\end{equation}

Finally, the ${\mu_{\mathcal{P}_j}}(q)$ is computed based on a normalized cross-entropy distance between $q_j$ and ($p^{\text{int}}_j$, $p^{mld}_j$) and triangular shape, according to the following equation:

\begin{equation}
\small
\mu_{\mathcal{P}_j}^{\text{mld/int}}(q_j) = max(0,0.5+\frac{\mp(\Delta H - \Delta\Bar{H})}{\sigma ^ \prime})
\label{equ:MFjointj}
\end{equation}

where, ($\Delta H = H({{P}_j^{\text{int}}},q_j) - H({{P}_j^{mld}},q_j)$) in which $H({{P}_j^{\text{int}}},q_j)$ and $H({{P}_j^{mld}},q_j)$ are the cross-entropy between the softmax activation of the attention weight distributions of intense-mild categories, in \autoref{equ:mld/intjointsdistibution}, and those of the current input sample computed in \autoref{equ:inputjointsdistibution}. $\Delta\Bar{H}$ is the average of these differences for all stored samples, and ${\sigma ^ \prime}$ defines the spread of the fuzzy set similar to $\sigma$ in \autoref{equ:MFintensityindex}. 

\textbf{Algorithm \ref{alg:membership_function}:} summarizes the step-by-step fuzzification procedure. The procedure consists of two average values of intensity score and difference cross-entropy which gets updated with every input video. It also includes two collections which stores the joint attention weights of for mild and intense categories which are classified by comparing the truth values of intensity scores, and by taking an average and softmax two probabilistic distributions are extracted for corresponding categories. These collections and their corresponding probabilistic distributions are dynamically appended and updated with every new input. The procedure gets the time frames and joints' attention weights as well as the key-point coordinates as input, computes the intensity score $I$ based on \autoref{equ:fuzzy_entropy_attn}, update its average value, maps it to fuzzy set $\mathcal{I}$ using triangular fuzzy membership function \autoref{equ:MFintensityindex}, and updates collections and their corresponding distributions. Next, it computes the $q$ \autoref{equ:inputjointsdistibution} and the cross-entropy between mild and intense distributions, update the cross-entropy average value, and maps the $q$ into fuzzy set $\mathcal{P}$ using fuzzy membership function \autoref{equ:MFjointj}. Finally, the procedure returns the truth values of $\mathcal{I}^{\text{mld}}$, $\mathcal{I}^{\text{int}}$, $\mathcal{P}^{\text{mld}}$, and $\mathcal{P}^{\text{int}}$.

\subsubsection{Fuzzy Rules and Inference for Final Indexing}
As mentioned in Section \ref{sec:::fuzzifier}, the final intensity index output is inferred based on fuzzy logic principles on the input sets ($\mathcal{I}$ and $\mathcal{P}_j \text{for all} j$). In specific, the input sets are passed through IF-THEN fuzzy logic rules, and then, by combining these rules the final output fuzzy sets are inferred which denote the intensity index of the performed action.

Initially, an intermediate output set is extracted by a linear combination of the following intermediate fuzzy rules, which we have for every set $\mathcal{P}_j$:

\begin{equation}
\small
 \text{R}^{\text{\text{mld/int}}}_j: \text{IF} \hspace{2mm} q_j \hspace{2mm} \text{is} \hspace{2mm} \mathcal{P}^{\text{mld/int}}_j \hspace{2mm} \text{THEN} \hspace{2mm} q \hspace{2mm} \text{is} \hspace{1mm} \mathcal{P}^{\text{mld/int}}, \hspace{1mm} \text{weight}=\alpha_j
 \label{equ:fuzzy_intermediate}
\end{equation}

where $\text{R}^{\text{\text{mld/int}}}_j$ is a set of two rules for every joint coordinates $j$, $\mathcal{P}^{\text{mld/int}}$ refers to $\mathcal{P}^{mld}$ and/or $\mathcal{P}^{\text{int}}$ members of the intermediate output fuzzy set $\mathcal{P}$ (i.e., $\mathcal{P} = \{\mathcal{P}^{mld},\mathcal{P}^{\text{int}}\}$) which denotes the aggregated categorization of joints' distribution into mild and/or intense. 
Each rule $\text{R}^{\text{\text{mld/int}}}_j$ refers to the corresponding joint's individual decision on the aggregated categorization whose role is weighted by $\alpha_j$. Next, we combine the inferences of these rules using the linear combination of their output fuzzy membership functions \cite{ross2005fuzzy}, to compute the overall membership function of the intermediate output set. This process is an adaptive filter as $\alpha_j$s are adaptively learned during the training session on the intensity indexing dataset \cite{mendel2017uncertain}.
Since the attention weights demonstrate the exclusive patterns of the action motions, the fuzzy rules will dynamically adapt to every category and index of actions. The fuzzy membership function of $\mathcal{P}^{\text{mld/int}}$ is formulated as follows:

\begin{equation}
 \mu_{\mathcal{P}}^{\text{mld/int}} (q) = \sum_j \alpha_j \mu_{{\mathcal{P}_j}\circ{\text{R}_j}}^{\text{mld/int}}(q_j) = \sum_j \alpha_j \mu_{\mathcal{P}_j}^{\text{mld/int}}(q_j)
 \label{equ:inter_fuzzy_membership}
\end{equation}

where, $\mu_{\mathcal{P}}^{\text{mld/int}} (q)$ is the intermediate fuzzy membership function, and $\mu_{{\mathcal{P}_j}\circ{\text{R}_j}}^{\text{mld/int}}(q_j)$ if the fuzzy membership function for every rule which measures the truth of the relation between $\mathcal{P}^{\text{mld/int}}$ and every joint's fuzzified distribution set $\mathcal{P}^{\text{mld/int}}_j$.

The final output inference of the intensity index is predicted using the intermediate output set and the fuzzified set of intensity score (i.e. $\mu_{\mathcal{I}}$), passed through the following final fuzzy inference rules:

\begin{equation}
 \text{R}^{\text{\text{mld}}}: \text{IF} \hspace{1mm} I \hspace{1mm} \text{is} \hspace{1mm} \mathcal{I}^{\text{mld}} \hspace{1mm} \text{AND} \hspace{1mm} q \hspace{1mm} \text{is} \hspace{1mm} \mathcal{P}^{\text{mld}} \hspace{1mm} \text{THEN} \hspace{1mm} y \hspace{1mm} \text{is} \hspace{1mm} Y^{\text{mld}}
 \label{equ:mild}
\end{equation}

\begin{equation}
 \text{R}^{\text{\text{int}}}: \hspace{1mm} \text{IF} \hspace{1mm} I \hspace{1mm} \text{is} \hspace{1mm} \mathcal{I}^{\text{int}} \hspace{2mm} \text{AND} \hspace{2mm} q \hspace{1mm} \text{is} \hspace{1mm} \mathcal{P}^{\text{int}} \hspace{2mm} \text{THEN} \hspace{2mm} y \hspace{1mm} \text{is} \hspace{1mm} Y^{\text{int}}
 \label{equ:intense}
\end{equation}

where $y$ is the final inference value which belongs to the final index set of $Y = \{Y^{mld}, Y^{int}\}$. The AND process is performed on fuzzy sets $\mathcal{I}^{\text{mld/int}}$ and $\mathcal{P}^{\text{mld/int}}$ using "AND-type" inference introduced by \cite{rutkowski2003flexible} which compromise linear combination of t-norm and s-norm of truth values $\mu_{\mathcal{I}}^{\text{mld/int}}(I), \mu_{\mathcal{P}}^{\text{mld/int}} (q)$, according to the following Equation:

\begin{equation}
\mu_Y^\text{mld/int} = \lambda \hspace{0.5mm}\text{t-norm}\hspace{1mm} +\hspace{1mm} \left( 1-\lambda \right) \hspace{0.5mm} \text{s-norm}
\end{equation}
where $\lambda$ parameter can be found in the process of learning subject to the constraint $0 < \lambda < 1$ along with $\langle \alpha_j \rangle$.

Finally, the intensity index is predicted by comparing $Y^{mld}$'s and  $Y^{int}$'s truth values, i.e. $\mu_Y^\text{mld}$ and $\mu_Y^\text{int}$. 

\begin{algorithm}[t!]
\footnotesize
{
    \caption{The summary of the fuzzy inference stage which gets the membership function of the fuzzified sets from  Algorithm \ref{alg:membership_function} and returns the final inference for intensity index. This procedure is performed for every input and return the final inference output fuzzy set whose membership function have the maximum truth value.}
    \label{alg:fuzzy_inference}
    \begin{algorithmic}

            \State $\text{R}^{\text{\text{mld/int}}}_j$: intermediate fuzzy inference rules for every joint $j$ (Eq. \ref{equ:fuzzy_intermediate})
            \State $\alpha_j$: represents the role of every joint in the final inference (Eq. \ref{equ:MFjointj}) 
            \State $\text{R}^{\text{\text{mld/int}}}$: final fuzzy inference rules (Eq. \ref{equ:mild} and Eq. \ref{equ:intense})
            \State $Y^{\text{mld/int}}$: final output sets for mild and intense indexes
            \For {every input $i$}
            \State $(\mu_{\mathcal{I}}^{\text{mld/int}}(I^i), \mu_{\mathcal{P}}^{\text{mld/int}} (q^i))$ = {FUZZIFIER}\big(${a^i}_t$, ${{a^{\prime}}^i}_{(j,t)}$, $x^i$\big)
            \Procedure{FUZZY-INFERENCE}{$(\mu_{\mathcal{I}}^{\text{mld/int}}(I^i), \mu_{\mathcal{P}}^{\text{mld/int}} (q^i))$}
            \For{every joint $j$} \Comment{{combination of intermediate inferences}}
            {
                \State $\mu_{\mathcal{P}}^{\text{mld}} (q)$  $+=$ $\alpha_j$ $\mu_{\mathcal{P}_j}^{\text{mld}}(q_j)$ (Eq. \ref{equ:inter_fuzzy_membership})  \Comment{{$R^{\text{mld}}_j$}}
                {
                \State $\mu_{\mathcal{P}}^{\text{int}} (q)$  $+=$ $\alpha_j$ $\mu_{\mathcal{P}_j}^{\text{int}}(q_j)$ (Eq. \ref{equ:inter_fuzzy_membership}) \Comment{{$R^{\text{int}}_j$ \hspace{0.5mm}}}
                {
                }}
            \EndFor   
            \State $\mu_Y^{mld} \gets$ AND-type$(\mathcal{I}^\text{mld},\mathcal{P}^\text{mld})$ \Comment{{$R^{\text{mld}}$\hspace{0.5mm}}}
            {
            \State $\mu_Y^{int} \gets$ AND-type$(\mathcal{I}^\text{int},\mathcal{P}^\text{int})$ \Comment{{$R^{\text{int}}$ \hspace{0.5mm}}}
            }
            {
            \State \textbf{return} {$\underset{Y \in \{Y^\text{mld},Y^\text{in}\}}{arg\hspace{1mm}max} \mu_Y\left(y\right)$}
            {
            \EndProcedure 
            \EndFor
            }}}
    \end{algorithmic}
}
\end{algorithm}

\textbf{Algorithm \ref{alg:fuzzy_inference}:} summarizes the process of our fuzzy inference system which outputs the final inference for intensity index based on input fuzzified sets $\mathcal{I^{\text{mld/int}}}$, $\mathcal{P^{\text{mld/int}}}$ from \autoref{sec:::fuzzifier}; along with intermediate and final fuzzy logic rules: $\text{R}^{\text{\text{mld/int}}}_j$ and $\text{R}^{\text{\text{mld/int}}}$ respectively; and combining their output fuzzy sets. The output sets of intermediate rules $\langle \text{R}^{\text{\text{mld/int}}}_j \rangle$ are combined using a linear combination method and $\langle \alpha_j \rangle$ weights which show the degree of belief to each rule. The output set of the final inference rules $\text{R}^{\text{\text{mld/int}}}$ are computed using AND-type of $\mathcal{I}^{\text{mld/int}}$ and $\mathcal{P}^{\text{mld/int}}$ \cite{rutkowski2003flexible}. Finally, the intensity index is inferred by comparing the truth values of the final output fuzzy sets $Y^{\text{mld}}$ and $Y^{\text{int}}$ corresponding to mild and intense indexes.

\subsection{Loss Function Update}

The ST-LSTM is initially trained on action recognition data of samples with similar intensity. However, actions performed with different intensities include different motion patterns. Consequently, the pre-trained ST-LSTM may pay attention the wrong joint coordinates once applied to the generated dataset which has samples of different intensities. Therefore, we add a penalty term to the loss function to enforce the model to pay attention to the intended joint coordinates by penalizing the wrong attention weights. In this regard, we use the cross-entropy as a distance between the input joints' distribution of \autoref{equ:inputjointsdistibution}, and those of mild and intense categories computed from \autoref{equ:mld/intjointsdistibution}. As such, the action recognition module of our methodology also adapts to the unique way a certain action-intensity is performed, e.g. `intense punching' vs. `mild punching.' This, addition of a penalty term, in turn, leads to the further adaptation of the kinetic fuzzy intensity score and of the output fuzzy rules. \autoref{equ:losspenalty} is the loss function of the LSTM model with the aforementioned penalty term added, enables the model to distinguish mild and intense actions, is given as:

\begin{equation}
{L}(y,\hat{y};p,q) = {{-{\sum\limits_{l}} {y}_{l} \log({\hat{y}}_{l})}-{\lambda}{\sum\limits_{j} {p}_{j} \log({q}_{j})}}
\label{equ:losspenalty}
\end{equation}

where, the first $log$-based term denotes cross-entropy which is used as a distance function between the real label and the computed softmax of the final output, i.e. $y$ and $\hat{y}$ respectively. $l$ is the index of the recognizable actions considered in the model. The penalty term is added through the Lagrange multiplier $\lambda$ \cite{vapnik1999overview, bouzary2019hybrid, pour2019data, begam2018flexible}, which increases with the number of input samples related to every action. The second $log$-based term is the cross-entropy penalty. $q$ denotes the input's distribution of attention weights over the joint coordinates, \autoref{equ:inputjointsdistibution},and $p$ denotes those of the mild and intense categories, \autoref{equ:mld/intjointsdistibution}. $j$ is the index of human key-points/joints coordinates. The loss function in \autoref{equ:losspenalty} enables the model to distinguish mild and intense actions. It improves action recognition accuracy when the action dataset includes mild and intense intensity indexes.

\section{Experiments}
\label{Experiments}
In this section, we elaborate upon: (i) our experimental setup, (ii) results of these experiments, (iii) the experimental discussions, in which we discuss the performance and limitations of our experiments.

\subsection{Experimental Setup}
\label{experimental_setup}
\subsubsection{Generated Dataset for Intensity Indexing} In order to evaluate the intensity indexing scheme and the whole integrated model, we generated an additional dataset of human actions with two intensity indexes: (i) \textit{intense}, and (ii) \textit{mild}. In part, our objective was to minimize the data requirements for our model, therefore, we employed a fuzzy system for action intensity indexing, which requires only a small amount of data. The choice of a fuzzy system allows for the use of the pre-existing SBU dataset which is small compared to UCF101 \cite{soomro2012ucf101} and NTU RGB+D \cite{shahroudy2016ntu} datasets. The SBU dataset contains 8 classes, 3 of which (exchanging object, hugging and shaking hands) cannot be differentiated as mild or intense due to the limitations of the model. Therefore, we extract the other 5 classes: \textit{approaching}, \textit{punching}, \textit{kicking}, \textit{hugging}, and \textit{pushing}. For each of these classes, we generated 100 intense and 100 mild videos. Therefore, the generated dataset includes 1000 samples. The classification of action intensity is subjective in nature, because for each person the perception of action intensity varies, and depends on their physical attributes (e.g. sex, age, height, BMI etc.) With that in mind, in our generated dataset annotation, we have requested students with similar physical attributes to perform the action with a certain intensity, and the classification was associated to the subject's own perception toward their performed actions. Generating more clusters (mild, medium, intense) for intensity indexing requires more data. As mentioned, we had to generate our own dataset with mild and intense indexes to evaluate the intensity indexing scheme. Therefore, to keep it simple, we decided to stick with just two clusters for the intensity indexing. For future work, more clusters can be added.

\subsubsection{Spatio-Temporal LSTM}
\label{sec:LSTM}
Firstly, we utilize the SBU Kinetic dataset \cite{yun2012two}, which is used for 3D classification of human key-point coordinates into an action class, to train the spatio-temporal LSTM \footnote{Because we used only 5 classes for the additional dataset generation, we stick to the same 5 classes for training the action recognition module.}. Each video in the SBU dataset is restricted to 2 people and each person has 15 joints targeted as key-point coordinates in each frame. Similar to Yun et al. \cite{yun2012two}, we apply a 5-fold cross validation scheme to evaluate the action recognition module. As shown in \autoref{table:results}, our ST-LSTM model with attention mechanism enhances the accuracy of the model of \cite{liu2016spatio} and achieves the state-of-the-art performance on SBU Kinetic dataset.

\begin{table}[t!]
\centering

\begin{tabular}{l|l}
\textbf{Method}                          & \textbf{Accuracy} \\\hline
Raw Skeleton \cite{yun2012two}                    & 49.7\%   \\
Joint Feature \cite{ji2014interactive}                  & 86.9\%   \\
CHARM \cite{li2015category}                          & 83.9\%  \\
Hierarchical RNN \cite{du2015hierarchical}                & 80.3\%  \\
Deep LSTM \cite{zhu2016co}                      & 86.3\%  \\
Deep LSTM + Co-occurrence \cite{zhu2016co}       & 87.4\%    \\ 
Clips + CNN + Concatenation \cite{ke2017new}     & 92.8\%  \\
Clips + CNN + Pooling  \cite{ke2017new}           & 92.2\%  \\
Clips + CNN + MTLN  \cite{ke2017new}             & 93.5\%  \\\hline
ST-LSTM (w/o Attention) \cite{liu2016spatio}                         & 88.6\%   \\ 
\text{ST-LSTM (w/o Attention) + Trust Gate \cite{liu2016spatio} }           & 93.3\%   \\ 
\textbf{ST-LSTM with Attention} & \textbf{94.2\%}  \\ \hline     
\end{tabular}
\caption{\small{Comparison of state-of-the-art action recognition models trained on SBU dataset. The results highlight the importance of the spatio-temporal attention mechanism which improves the accuracy of the ST-LSTM.}}
\label{table:results}
\end{table}


\subsection{Experimental Results}
\label{sec:results}
\subsubsection{Action Recognition}

We use the trained Spatio-Temporal LSTM module and fine tune it with the additional dataset which consists of actions performed with various intensity. Since the mild and intense actions are performed differently, in terms of motional patterns, the accuracy of the ST-LSTM module drops significantly, up to 19\%. Therefore, we dynamically update the LSTM model with the results of the fuzzy inference system, according to \autoref{equ:losspenalty}, and re-evaluate it on the generated dataset to see how the action recognition module's accuracy would be influenced by the integration of the LSTM and intensity indexing modules. \autoref{table:AR_ourdata} depicts the re-evaluation results on our additional generated dataset showing the average $2.75\%$ decrease in the overall accuracy. 

\begin{table}[b!]
\centering
\begin{tabular}{l|l}
\textbf{Action}      & \textbf{Accuracy} \\ \hline
Punching & 93.8\% \\ 
Kicking & 92.7\% \\ 
Pushing & 93.3\% \\ 
Hugging & 88.7\% \\ 
Approaching & 89.1\% \\ \hline
\textbf{Overall} & \textbf{91.2\%} \\ \hline
\end{tabular}
\caption{\small{The re-evaluation results of our action recognition model on the generated dataset. The loss function has been updated according to \autoref{equ:losspenalty} which help the model performance not to decrease significantly.}}
\label{table:AR_ourdata}
\end{table}

\subsubsection{Intensity Indexing}
We use the additional generated dataset to evaluate the performance of our intensity indexing methodology. \autoref{table:intensity_index} shows the action intensity indexing performance of our model on the generated dataset. We have considered the fuzzy inference rules in \autoref{equ:intense} and \autoref{equ:mild} separately, to measure the F1 score and reported the averaged results. Due to the strictness of the fuzzy inference rules, the precision of the intensity indexing is comparably higher than other metrics. By using both fuzzy rules jointly, we would reach higher precision; however, there would be samples that are not classified as \textit{intense} or \textit{mild}.

Actions like hugging and approaching are tough to distinguish between intense and mild. The fuzzy module of the proposed system takes input from the attention weights generated by the spatio-temporal LSTM. These attention weights are of two type: one over the time frame, and another over the key-point coordinates in every frame.  Key-point coordinates for approaching and hugging do not differ by much for intense and mild classes, resulting in similar attention weights for both intensity indexes, which makes the model drop in accuracy as seen in \autoref{table:intensity_index}

We further compare our methodology with multi-task learning baselines implemented on top of our ST-LSTM to comprehend the role of the fuzzy kinetic analysis. The evaluation results of \autoref{tab:baseline_comparison} demonstrates the significance of the kinetic fuzzy intensity analysis and indexing modules of our methodology. Similar to the evaluation scheme in action recognition module, we use 5-fold cross validation to evaluate the intensity indexing algorithm for each action class. For evaluation of a model with limited data samples, k-fold cross validation process is used. In our 5-fold cross validation we use 4 folds for training and the remaining 1-fold for testing. As for the baselines, we use the attention weights as the input features to the SVM and DNN. There are two kind of attention mechanism applied on the time frames: one over every frame in the video and another over the key-point coordinates in every frame. These attention mechanisms generate the attention weights which are the input features to the SVM and DNN. The use of SVM, regression or fuzzy modules is to classify the action intensity as intense or mild whereas the ST-LSTM recognizes the action and together the output is the action plus its nature in term of intensity. The use of regression and SVM was to compare their performance with fuzzy for the task of intensity indexing from the attention weights generated by the ST-LSTM.



\begin{figure*}
\begin{subfigure}{.5\linewidth}
  \centering
  \includegraphics[width=\linewidth]{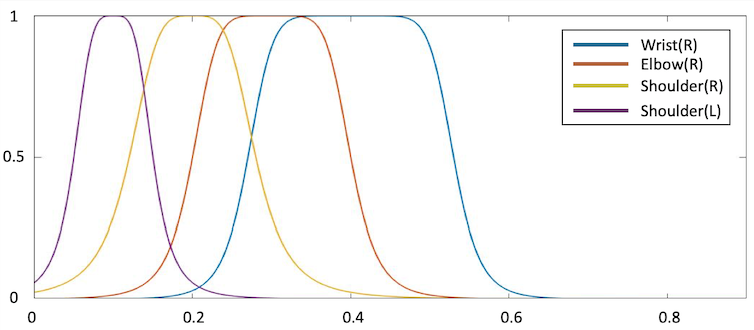}
  \caption{Punching Intense}
  \label{fig:bell_1}
\end{subfigure}%
\begin{subfigure}{.5\linewidth}
  \centering
  \includegraphics[width=\linewidth]{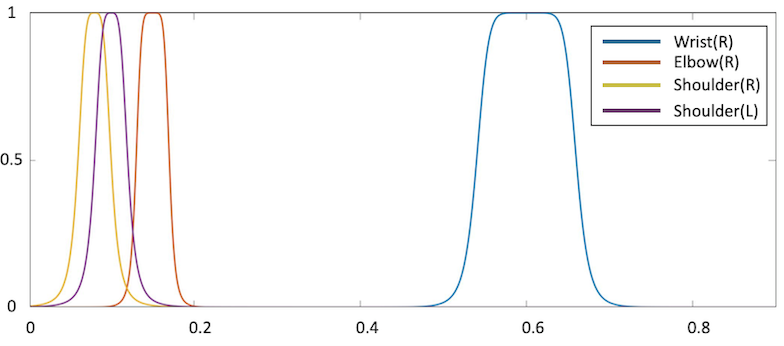}
  \caption{Punching Mild}
  \label{fig:bell_2}
  \end{subfigure}
  \begin{subfigure}{.5\linewidth}
  \centering
  \includegraphics[width=\linewidth]{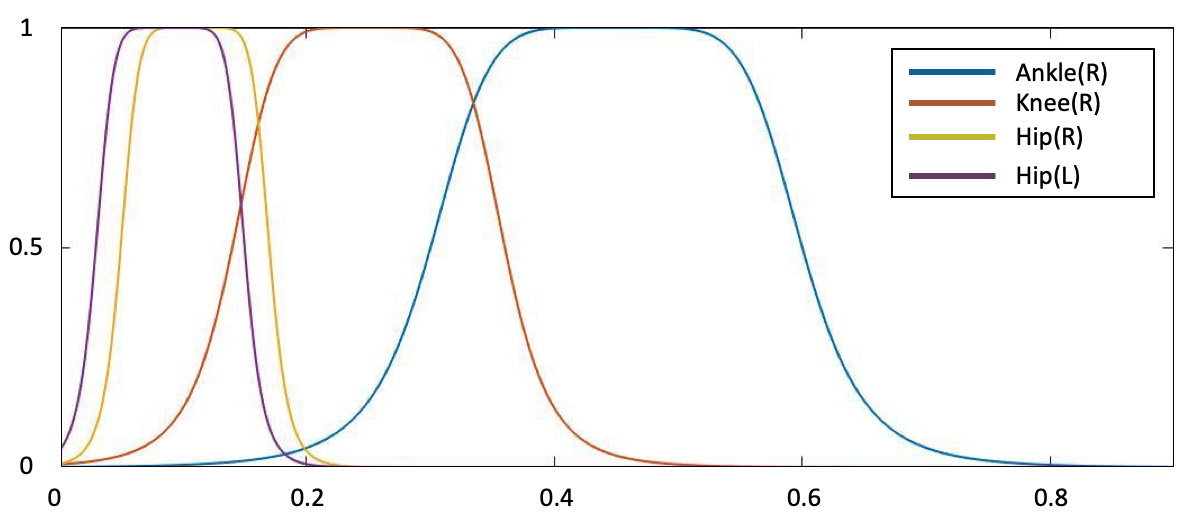}
  \caption{Kicking Intense}
  \label{fig:bell_3}
\end{subfigure}%
\begin{subfigure}{.5\linewidth}
  \centering
  \includegraphics[width=\linewidth]{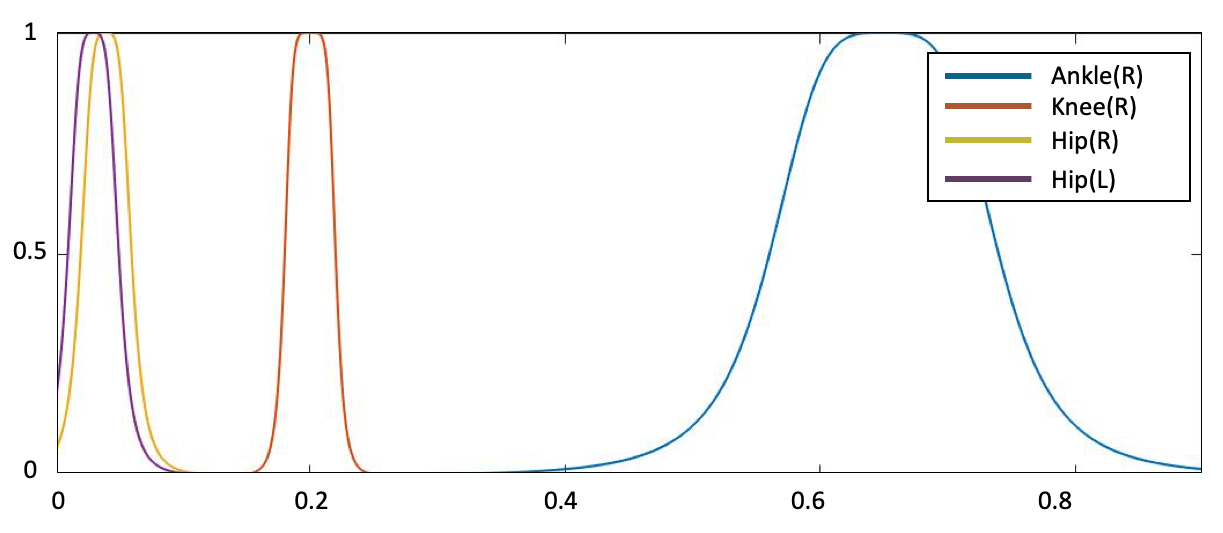}
  \caption{Kicking Mild}
  \label{fig:bell_4}
  \end{subfigure}
\caption{\small{We fit the generalized bell membership function to these distributions by assigning a membership score of 1 if the detected index is intense, and 0.5 if the distribution is closer to the intense category but the final intensity index is not detectable, i.e. intensity score is below the threshold. The attention weights are multiplied by the temporal attention of the time frames.}}
\label{fig:attention graph}
\end{figure*}

\begin{table}[b!]
\centering
\begin{tabular}{l|l|l|l|l}
\multicolumn{5}{c}{}          \\ 
\textbf{Action}      & \textbf{Accuracy} & \textbf{F1 Score} & \textbf{Precision} & \textbf{Recall} \\\hline
Punching    & 95\%     & 94.7\%   & 100\%     & 90\%   \\
Kicking     & 94\%     & 93.6\%   & 100\%     & 88\%   \\
Pushing     & 90\%     & 89.4\%   & 95.5\%    & 84\%   \\
Hugging     & 89\%     & 88.7\%   & 91.5\%    & 86\%   \\
Approaching & 83\%     & 82.5\%   & 85.1\%    & 80\%   \\\hline
\textbf{Average}     & \textbf{89.16\%}  & \textbf{88.76\%}  & \textbf{92.91\%}   & \textbf{85\%} \\ \hline
\end{tabular}
\caption{\small{Experimental results of the intensity indexing performed on the generated dataset. There is a subtle difference between the mild and intense variation for hugging and approaching class which causes the accuracy to drop.}}
\label{table:intensity_index}
\end{table}

\subsection{Experimental Discussions}

\subsubsection{Joints' Distributions in Intense vs. Mild Indexes}

As mentioned in \autoref{sec:::fuzzifier}, our model dynamically learns the motion of every index of each action category through the weighted distribution of joints corresponding to the attention weights of the LSTM module. \autoref{fig:attention graph} shows these distributions for actions of punching and kicking with mild and intense indexes. We fit a generalized bell function \cite{jang1993anfis} to these distributions by assigning 1.0 to them if the intensity score is above the average $\Bar{I}$ and the cross entropy of intense distribution is less than the mild distribution, 0.5 if the intensity score is less than the average but the cross entropy with the intense distribution is still less, and 0 otherwise. The   \autoref{fig:attention graph} shows, on the aggregated level, the distinct distribution of these weights for intense-mild actions. In \autoref{fig:attention graph}, we fit the generalized bell membership function to joints' attention weights extracted from the generated dataset. It illustrates the difference between mild and intense actions in terms of joints' movement and the weight by which the action recognition module is attending to them. As shown in the figure, the distribution of these attention weights for the intense actions tend to have higher variance whereas in the mild actions they are rather dense around the average value. In addition, while the intense actions tend to have a greater number of joints with significant corresponding attention weights, as for the mild actions the attention weight of only one joint have significant value and the rest have trivial values below 0.2. This stand to the reason that our model generates fuzzy membership values of intensity indexes for every joint's motional patterns and utilizes it for final indexing inference.

\subsubsection{Restrictions}
\begin{figure}[b!]
\centering
\begin{subfigure}{.3\linewidth}
  \centering
  \includegraphics[width=0.6\linewidth]{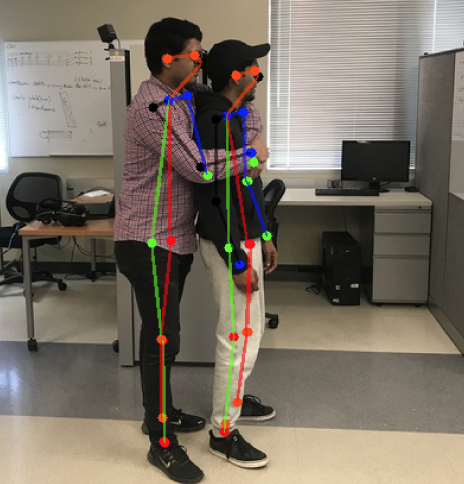}
  \caption{\small{False Negative}}
  \label{fig:false_negative}
\end{subfigure}%
\begin{subfigure}{.3\linewidth}
  \centering
  \includegraphics[width=0.6\linewidth]{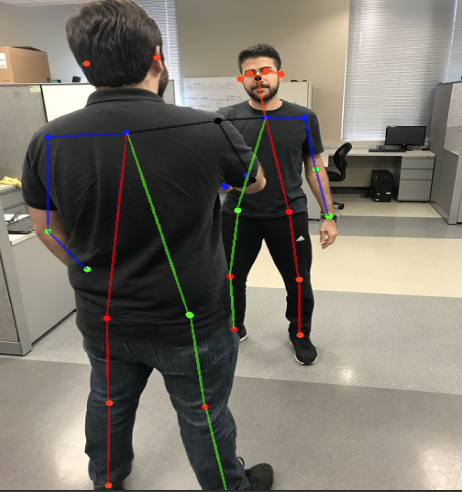}
  \caption{\small{Camera Angle}}
  \label{fig:camera_angle}
\end{subfigure}
\begin{subfigure}{0.6\linewidth}
  \centering
  \includegraphics[width=0.6\linewidth]{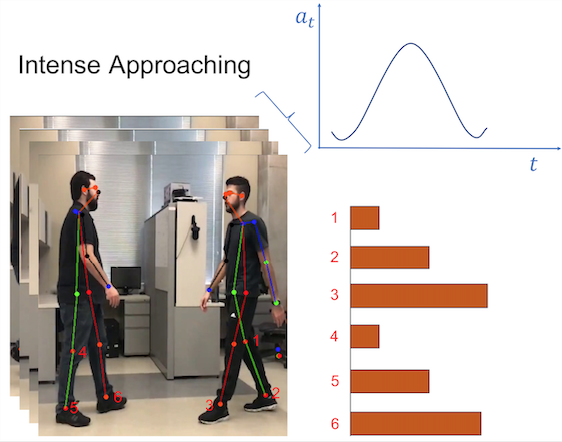}
  \caption{\small{False Positive}}
  \label{fig:false_positive}
\end{subfigure}
\caption{\small{Example on false negative, incorrect prediction and false positive due to poor camera angle on test videos by the Spatio-Temporal LSTM model.}}
\label{fig:action_recognition}
\end{figure}

In this subsection, we discuss the restrictions of our fuzzy recurrent attention model in detail, i.e. false positives, false negatives and explainable cases of misclassification. The action recognition module has a separate pre-processing unit whose function is not to achieve a contextual understanding from the scene but merely to extract the human key-point coordinates. Therefore, the model gets a limited understanding from the scene which does not include details such as camera angle, subjects' directions, etc. \autoref{fig:action_recognition} demonstrates such weaknesses of the action recognition module by providing concrete examples of the samples which have been misclassified due to the limited contextual understanding of the scene. 

\begin{table}[t!]
\centering
\begin{tabular}{l|l} 
\textbf{Model} & \textbf{Accuracy} \\ \hline
ST-LSTM + Regression & 76.4\% \\
ST-LSTM + SVM (Support Vector Machine) & 79.4\% \\
ST-LSTM + DNN (Dense Neural Network) & 82.1\% \\ \hline
\textbf{ST-LSTM + Fuzzy Kinetic Analysis} & \textbf{89.2\%} \\ \hline
\end{tabular}
\caption{\small{Baseline Model Comparison. Experimental Results on the generated dataset show the fuzzy outperforms multi-task learning baselines implemented using Regression, SVM, DNN on top of ST-LSTM.}}
\label{tab:baseline_comparison}
\end{table}

In \autoref{fig:false_negative}, the action performed is hugging but the model failed to predict the performed action. Since our model is trained using the human key-point location and predicts using the same, the model failed to predict the action as hugging. If both the subjects were facing one another, the prediction could have been correct. In \autoref{fig:camera_angle}, the two humans are shaking hands but since the camera angle is not sideways the key-point location do not indicate the action of shaking hands and because of the wrong camera angle, the model fails to recognize the action performed. As for the intensity indexing unit, the temporal distribution of attention weights over the time frames does not distinguish between the key points of the subjects performing the action, while the speed of action, in \autoref{equ:fuzzy_1}, refers to actions which are performed by every individual subject. As such, the model might malfunction in cases where two subjects are performing the action that previously had been performed by a single subject. As a concrete example, in \autoref{fig:false_positive}, the two humans are walking slowly toward each other, but as the key point of both of the humans are approaching fast, the temporal attention over the time frames indicates it to be an intense action, whereas, in reality the action performed was of mild.

\section{Conclusion}
\label {Conclusion}
In this paper, we incorporate fuzzy logic based inference into neural-based action recognition systems to tackle the task of intensity indexing from video inputs. We propose a hybrid model of fuzzy logic in conjunction with a spatio-temporal LSTM network equipped with an attention mechanism, kinetics and fuzzy logic concepts as well as a fuzzy inference system. Our research shows that the implemented fuzzy logic component is able to handle the uncertainty inherent to interpreting action intensity from a video. The model was able to achieve a testing accuracy of 89.16\% on our generated dataset for the task of intensity indexing. We were also able to determine the dynamic fuzzy logic rules to detect the intensity index for different action classes. 

In the future, to help detection of aggressive and bullying behavior, we suggest using an enhanced version of our fuzzy recurrent attention model to perform action recognition with more classes of actions, with one or multiple objects which are intra-related. Enhanced version refers to the future work where we intend to add more sophistication to the model by adding more action classes and intensity indexes compared to the two indexes used in the current version: intense and mild. Coupled with these modifications, we also aim to improve the adaptive learning of the model, thus making it a more enhanced version of this base model.



\bibliographystyle{IEEEtran}
\bibliography{main}


\end{document}